# Evaluating Real-time Anomaly Detection Algorithms – the Numenta Anomaly Benchmark

Alexander Lavin
Numenta, Inc. Redwood City, CA
alavin@numenta.com

Subutai Ahmad
Numenta, Inc. Redwood City, CA
sahmad@numenta.com

October 9, 2015

# Evaluating Real-time Anomaly Detection Algorithms – the Numenta Anomaly Benchmark

Alexander Lavin
Numenta, Inc. Redwood City, CA
alavin@numenta.com

Subutai Ahmad
Numenta, Inc. Redwood City, CA
sahmad@numenta.com

*Abstract*

*Much of the world's data is streaming, time-series data, where anomalies give significant information in critical situations; examples abound in domains such as finance, IT, security, medical, and energy. Yet detecting anomalies in streaming data is a difficult task, requiring detectors to process data in real-time, not batches, and learn while simultaneously making predictions. There are no benchmarks to adequately test and score the efficacy of real-time anomaly detectors. Here we propose the Numenta Anomaly Benchmark (NAB), which attempts to provide a controlled and repeatable environment of open-source tools to test and measure anomaly detection algorithms on streaming data. The perfect detector would detect all anomalies as soon as possible, trigger no false alarms, work with real-world time-series data across a variety of domains, and automatically adapt to changing statistics. Rewarding these characteristics is formalized in NAB, using a scoring algorithm designed for streaming data. NAB evaluates detectors on a benchmark dataset with labeled, real-world time-series data. We present these components, and give results and analyses for several open source, commercially-used algorithms. The goal for NAB is to provide a standard, open source framework with which the research community can compare and evaluate different algorithms for detecting anomalies in streaming data.*

## I. Introduction

With the rapid rise in real-time data sources the detection of anomalies in streaming data is becoming increasingly important. Use cases such as preventative maintenance, fraud prevention, fault detection, and monitoring can be found throughout numerous industries such as finance, IT, security, medical, energy, e-commerce, and social media. Anomaly detection is notoriously difficult to benchmark and compare [1, 2]. In addition, real-time applications impose their own unique constraints and challenges that must be considered. The goal of this paper is to introduce the Numenta Anomaly Benchmark (NAB), a rigorous new benchmark and source code for evaluating real-time anomaly detection algorithms.

Anomaly detection in real-world streaming applications is challenging. The detector must process data and output a decision in real-time, rather than making many passes through batches of files. In most scenarios the number of sensor streams is large and there is little opportunity for human, let alone expert, intervention. As such, operating in an unsupervised, automated fashion (e.g. without manual parameter tweaking) is often a necessity. As part of this automation, the detectors should continue to learn and adapt to changing statistics while simultaneously making predictions. The real goal is often prevention, rather than detection, so it is desirable to detect anomalies as early as possible, giving actionable information ideally well before a catastrophic failure.

Benchmarks designed for static datasets do not adequately capture the requirements of real-time applications. For example, scoring with standard classification metrics such as precision and recall do not suffice because they fail to reflect the value of early detection. An artificial separation into training and test sets does not properly capture a streaming scenario nor does it properly evaluate a continuously learning algorithm. The NAB methodology and scoring rules (described below) are designed with such criteria in mind. Through experience with customers and researchers we also discovered it would be beneficial for the industry to include real-world labeled data from multiple domains. Such data is rare and valuable, and NAB attempts to incorporate such a dataset as part of the benchmark. There exist two other time-series data corpuses intended for real-time anomaly detection: the UC-Irvine dataset [3] and a recently released dataset from Yahoo Labs [4]. Neither of these include a scoring system but their data could eventually be incorporated into NAB.

NAB attempts to provide a controlled and repeatable environment of tools to test and measure different anomaly detection algorithms on streaming data. We include in this paper an initial evaluation of four different real-time algorithms. At Numenta we have developed an anomaly detection algorithm based on Hierarchical Temporal Memory (HTM). HTM is a continuous learning system derived from theory of the neocortex [5] and is well suited for real-time applications. The algorithm has proven useful in applications such as monitoring server data, geospatial tracking, stock trading metrics, and social media [6]. We also include comparative results with open source anomaly detection algorithms from Etsy Skyline [7], a popular open source algorithm, and two from Twitter [8]. There are, of course, many algorithms we have not directly tested [2, 9-14]. It is our hope that eventually a wide assortment of algorithms will be independently evaluated and results reported in a manner that is objectively comparable.

In the next section we discuss the two main components of NAB: the scoring system and dataset. We then discuss and analyze NAB scoring results for the above algorithms.

## II. NUMENTA ANOMALY BENCHMARK

NAB aims to represent the variety of anomalous data and the associated challenges detectors face in real-world streaming applications. We define anomalies in a data stream to be patterns that do not conform to past patterns of behavior for the stream. This definition encompasses both point anomalies (or spatial anomalies) as well as temporal anomalies. For example, a spiking point anomaly occurs when a single data point extends well above or below the expected range. Streaming data commonly also contains temporal anomalies, such as a change in the frequency, sudden erratic behavior of a metric, or other temporal deviations. Anomalies are defined with respect to past behavior. This means a new behavior can be anomalous at first but ceases to be anomalous if it persists; i.e. a new normal pattern is established. Fig. 1 shows a few representative anomalies taken from the NAB dataset.

In the next two sections we discuss both the NAB dataset and scoring system, and the qualities that make them ideal for evaluating real-world anomaly detection algorithms.

### A. Benchmark Dataset

In the current version of NAB we focus on time-series data where each row contains a time stamp plus a single scalar value. The requirements are then to (i) include all types of streaming data anomalies, (ii) include a variety of data metrics, and (iii) present common challenges such as noise and establishing new normal patterns.

Anomalous patterns differ significantly across applications. A one-second latency in periodic EKG data could be a significant fluctuation, but the same pattern in stock trading volume may be meaningless. It is thus important for the NAB dataset to include metrics across a variety of domains and applications. The data currently in the NAB corpus represents a variety of metrics ranging from IT metrics such as network utilization to sensors on industrial machines to social media chatter. We also include some artificially-generated data files that test anomalous behaviors not yet represented in the corpus's real data, as well as several data files without any anomalies. The current NAB dataset contains 58 data files, each with 1000-22,000 data instances, for a total of 365,551 data points.

The NAB dataset is labeled by hand, following a meticulous, documented procedure. Labelers must adhere to a set of rules when inspecting data files for anomalies, and a label-combining algorithm formalizes agreement into ground truth labels. The process is designed to mitigate human error as much as possible.[1] In addition a smooth scoring function (described below) ensures that small labeling errors will not cause large changes in reported scores.

It is often prohibitively expensive to collect an accurately labeled set of anomalous data instances that covers all types of anomalous behavior [2]. A key element of the NAB dataset is the inclusion of real-world data with anomalies for which we know the causes. We propose the NAB dataset as a quality collection of time-series data with labeled anomalies, and that it is well suited to be a standard benchmark for streaming applications.

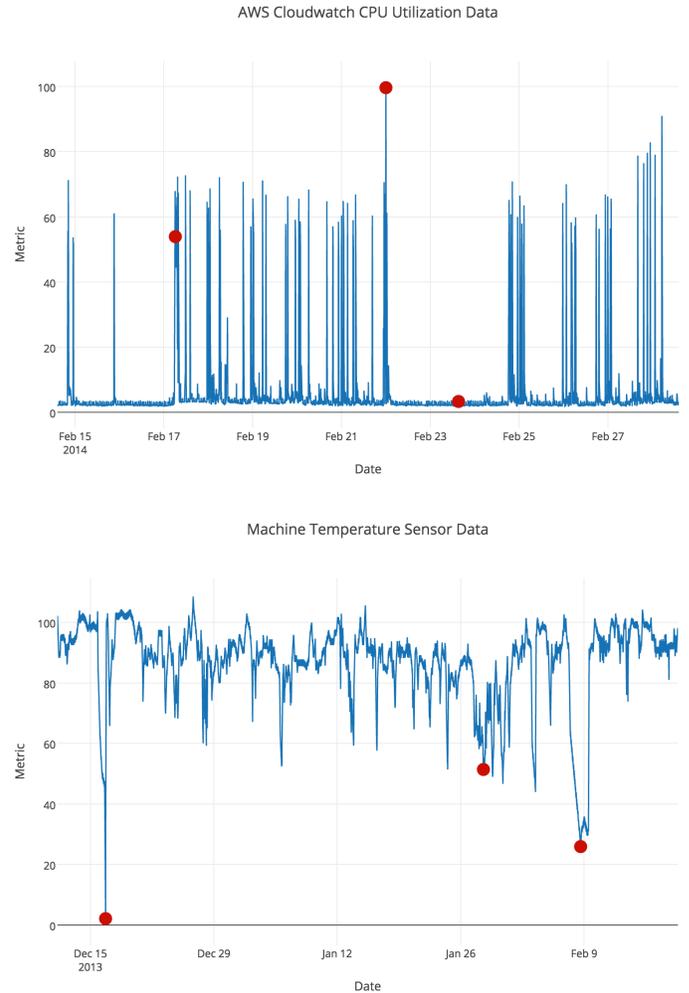

Fig. 1. Two representative examples of real data streams from the NAB dataset. Anomalies are labeled with red circles. The first anomaly in the top figure is subtle and challenging. The spiking behavior does not return to the baseline as expected, and this is soon the new normal pattern. The second anomaly is a simple spike anomaly after which the system returns to previous patterns. The third anomaly identifies a long period inconcsitent with the normal spiking pattern. The bottom figure shows temperature sensor data from an internal component of a large, expensive, industrial machine. The first anomaly was a planned shutdown. The third anomaly is a catastrophic system failure. The second anomaly, a subtle but observable change in the behavior, indicated the actual onset of the problem that led to the eventual system failure.

### B. Scoring Real-Time Anomaly Detectors

In NAB an anomaly detector accepts data input and outputs instances which it deems to be anomalous. The NAB scoring system formalizes a set of rules to determine the overall quality of anomaly detection. We define the requirements of the ideal, real-world anomaly detector as follows:

i. detects all anomalies present in the streaming data
ii. detects anomalies as soon as possible, ideally before the anomaly becomes visible to a human
iii. triggers no false alarms (no false positives)
iv. works with real time data (no look ahead)

---

[1] The full labeling process and rules can be found in the NAB wiki, along with the label-combining source code, in the NAB repo [15].

v. is fully automated across all datasets (any data specific parameter tuning must be done online without human intervention)

The NAB scoring algorithm aims to reward these characteristics. There are three key aspects of scoring in NAB: anomaly windows, the scoring function, and application profiles. These are described in more detail below.

Traditional scoring methods, such as precision and recall, don't suffice because they don't effectively test anomaly detection algorithms for real-time use. For example, they do not incorporate time and do not reward early detection. Therefore, the standard classification metrics – true positive (TP), false positive (FP), true negative (TN), and false negative (FN) are not applicable for evaluating algorithms for the above requirements.

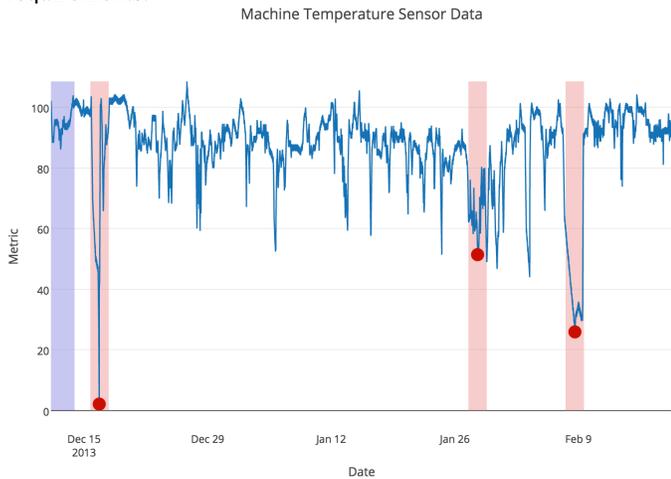

Fig. 2. Shaded red regions represent the anomaly windows for this data file. The shaded purple region is the first 15% of the data file, representing the *probationary period*. During this period the detector is allowed to learn the data patterns without being tested.

To promote early detection NAB defines *anomaly windows*. Each window represents a range of data points that is centered around a ground truth anomaly label. Fig. 2 shows an example using the data from Fig 1. A *scoring function* (described in more detail below) uses these windows to identify and weight true positives, false positives, and false negatives. If there are multiple detections within a window, the earliest detection is given credit and counted as a true positive. Additional positive detections within the window are ignored. The sigmoidal scoring function gives higher positive scores to true positive detections earlier in a window and negative scores to detections outside the window (i.e. the false positives). These properties are illustrated in Fig. 3 with an example.

How large should the windows be? The earlier a detector can reliably identify anomalies the better, implying these windows should be as large as possible. The tradeoff with extremely large windows is that random or unreliable detections would be regularly reinforced. Using the underlying assumption that true anomalies are rare, we define anomaly window length to be 10% the length of a data file, divided by the number of anomalies in the given file. This technique allows us to provide a generous window for early detections and also allow the detector to get partial credit if detections are soon after the ground truth anomaly. 10% is a convenient number but note the exact number is not critical. We tested a range of window sizes (between 5% and 20%) and found that, partly due to the scaled scoring function, the end score was not sensitive to this percentage.

Different applications may place different emphases as to the relative importance of true positives vs. false negatives and false positives. For example, the graph in Fig. 2 represents an expensive industrial machine that one may find in a manufacturing plant. A false negative leading to machine failure in a factory can lead to production outages and be extremely expensive. A false positive on the other hand might require a technician to inspect the data more closely. As such the cost of a false negative is far higher than the cost of a false positive. Alternatively, an application monitoring the statuses of individual servers in a datacenter might be sensitive to the number of false positives and be fine with the occasional missed anomaly since most server clusters are relatively fault tolerant.

To gauge how algorithms operate within these different application scenarios, NAB introduces the notion of *application profiles*. For TPs, FPs, FNs, and TNs, NAB applies different relative weights associated with each profile to obtain a separate score per profile.

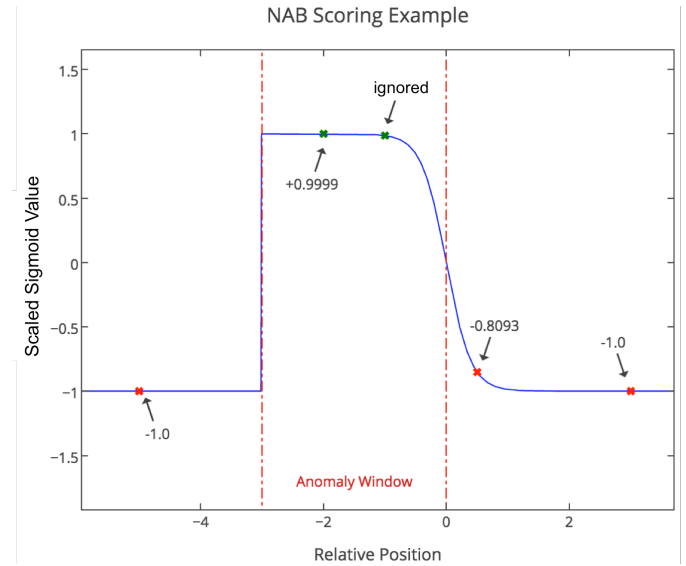

Fig. 3. Scoring example for a sample anomaly window, where the values represent the scaled sigmoid function, the second term in Eq. (1). The first point is an FP preceding the anomaly window (red dashed lines) and contributes -1.0 to the score. Within the window we see two detections, and only count the earliest TP for the score. There are two FPs after the window. The first is less detrimental because it is close to the window, and the second yields -1.0 because it's too far after the window to be associated with the true anomaly. TNs make no score contributions. The scaled sigmoid values are multiplied by the relevant application profile weight, as shown in Eq. (1), the NAB score for this example would calculate as: $-1.0A_{FP} + 0.9999A_{TP} - 0.8093A_{FP} - 1.0A_{FP}$. With the standard application profile this would result in a total score of 0.6909.

NAB includes three different application profiles: standard, reward low FPs, and reward low FNs. The standard profile assigns TPs, FPs, and FNs with relative weights (tied to the window size) such that random detections made 10% of the time would get a zero final score on average. The latter two profiles accredit greater penalties for FPs and FNs, respectively. These two profiles are somewhat arbitrary but designed to be illustrative of algorithm behavior. The NAB codebase itself is designed such that the user can easily tweak the relative weights and re-score all algorithms. The application profiles thus help evaluate the sensitivity of detectors to specific applications criteria.

The combination of anomaly windows, a smooth temporal scoring function (details in the next section), and the introduction of application profiles allows researchers to evaluate online anomaly detector implementations against the requirements of the ideal detector. Specifically the overall NAB scoring system evaluates real-time performance, prefers earlier detection of anomalies, penalizes "spam" (i.e. FPs), and provides realistic costs for the standard classification evaluation metrics TP, FP, TN, and FN.

### C. Computing NAB Score: Details

The final NAB score for a given algorithm and a given application profile is computed as follows. Let $A$ be the application profile under consideration, with $A_{TP}, A_{FP}, A_{FN}, A_{TN}$ the corresponding weights for true positives, false positives, etc. These weights are bounded $0 \leq A_{TP}, A_{TN} \leq 1$ and $-1 \leq A_{FP}, A_{FN} \leq 0$. Let $D$ be the set of data files and let $Y_d$ be the set of data instances detected as anomalies for datafile $d$. (As discussed earlier, we remove redundant detections: if an algorithm produces multiple detections within the anomaly window, we retain only the earliest one.) The number of windows with zero detections in this data file is the number of false negatives, represented by $f_d$.

The following scaled sigmoidal scoring function defines the weight of individual detections given an anomaly window and the relative position of each detection:

$$\sigma^A(y) = (A_{TP} - A_{FP})\left(\frac{1}{1+e^{5y}}\right) - 1 \tag{1}$$

In Eq. (1), $y$ is the relative position of the detection within the anomaly window. The parameters of Eq. (1) are set such that the right end of the window evaluates to $\sigma(y = 0.0) = 0$ (see Fig. 3), and it yields a max and min of $A_{TP}$ and $A_{FP}$, respectively. Every detection outside the window is counted as a false positive and given a scaled negative score relative to the preceding window. The function is designed such that detections slightly after the window contribute less negative scores than detections well after the window. Missing a window completely is counted as a false negative and assigned a score of $A_{FN}$.

The raw score for a data file is the sum of the scores from individual detections plus the impact of missing any windows:

$$S_d^A = \left(\sum_{y \in Y_d} \sigma^A(y)\right) + A_{FN} f_d \tag{2}$$

Eq. (2) accumulates the weighted score for each true positive and false positive, and detriments the total score with a weighted count of all the false negatives. The benchmark raw score for a given algorithm is simply the sum of the raw scores over all the data files in the corpus:

$$S^A = \sum_{d \in D} S_d^A \tag{3}$$

The final reported score is a normalized NAB score computed as follows:

$$S_{NAB}^A = 100 \cdot \frac{S^A - S_{null}^A}{S_{perfect}^A - S_{null}^A} \tag{4}$$

Here we scale the score based on the raw scores of a "perfect" detector (one that outputs all true positives and no false positives) and a "null" detector (one that outputs no anomaly detections). It follows from Eq. (4) that the maximum (normalized) score a detector can achieve on NAB is 100, and an algorithm making no detections will score 0.

### D. Other NAB Details

NAB is most valuable as a community tool, benefitting researchers in academia and industry. In building NAB we have been collaborating with the community, and welcome more contributions of data and additional online anomaly detection algorithms. To this end, NAB is a completely open source code base released under the permissive MIT License [15]. It follows a versioning protocol and public issue tracking, allowing changes and dataset additions to be clearly discussed and communicated. We post scores from contributed algorithms on the NAB scoreboard [16], which reflects a specific NAB version number – NAB v1.0 at the time of paper submission.[1] The codebase is modular and designed to make it easy to test additional algorithms (regardless of programming language), adjust application profiles, and even test custom labeled datafiles.

### III. ALGORITHMS TESTED

It is our hope that over time the NAB scoreboard will reflect results from a large number of algorithms. In this paper we report initial NAB results using four open source and commercially used algorithms, plus some control detectors. The four primary algorithms are the Numenta HTM anomaly detector, Etsy Skyline, and two Twitter algorithms, AnomalyDetectionTs and AnomalyDetectionVec. We also use some simple control detectors as baselines. Each of these are briefly described below.

The HTM detector (developed by Numenta and the authors) is based on Hierarchical Temporal Memory (HTM), a machine intelligence technology inspired by the structure of the neocortex [17, 18, 19]. Given a real-time data stream $\ldots, x_{t-2}, x_{t-1}, x_t, x_{t+1}, x_{t+2}, \ldots$, the algorithm models the temporal sequences in that stream. At any point $t$ the HTM makes multiple predictions for $x_{t+1}$. At time $t+1$ these predictions are compared with the actual values to determine an instantaneous anomaly score between 0 and 1. If any of the predictions are significantly different from $x_{t+1}$ the anomaly score will be 1. Conversely, if one of the predictions is exactly equal to $x_{t+1}$ the anomaly score will be 0. The system maintains the mean and variance of the recent distribution of anomaly scores. At every time step it outputs the likelihood that the current anomaly score is from the respective normal distribution. The anomaly likelihood score is thresholded to detect the final anomaly.

---
[1] Contributor details can be found in the NAB wiki.

The Numenta algorithm has several features that make it suitable for real world streaming data. It can deal with both predictable and highly unpredictable data – the final score is the deviation from the typical level of predictability for that stream. The temporal model makes multiple predictions and can thus deal with branching temporal sequences. The algorithm is a continuously learning algorithm, so changes in the statistics of the data are automatically handled without additional retraining. Finally, the system is robust to parameter settings. Thus it is able to handle a wide range of datasets without manual parameter tweaking. The code has been used in commercial applications and is freely available under an AGPL license in Python and C++ [20].

Skyline is a real-time anomaly detection system [7] originally developed by Etsy.com for monitoring its high traffic web site. The algorithm employs a mixture of experts approach. It incorporates a set of simple detectors plus a voting scheme to output the final anomaly score. The detectors include deviation from moving average, deviation from a least squares estimate, deviation from a histogram of past values, etc. Like the Numenta algorithm, Skyline is well suited for analyzing streaming data. The internal estimates are continually adjusted and, due to the mixture of simple experts, it is relatively robust across a wide range of applications. The code is open source and has been tested in commercial settings.

Twitter recently released two versions of a real-time anomaly detection algorithm. It uses a combination of statistical techniques to robustly detect outliers. The Generalized ESD test [21] is combined with robust statistical metrics, and piecewise approximation is used to detect long term trends. One version of the algorithm AnomalyDetectionVec is intended to be more general and detect anomalies in data without timestamps but requires manually tuning of the periodicity. A second version of the algorithm, AnomalyDetectionTs, exploits timestamps to detect periodicity and can detect both short-term (intra-day) and long-term (inter-day) anomalies. Unfortunately AnomalyDetectionTs was unable to calculate the necessary period parameters for some of the NAB data files, and thus we cannot include it in the Table 1 results. It is worth noting that for the data files ADTs successfully ran, it was outperformed by ADVec. Both algorithms have been used in commercial settings and the R code is available as open source [22].

In addition to the above core set of detectors we use three control detectors. A "null" detector outputs a constant anomaly score of 0.5 for all data instances. A "perfect" detector is an oracle that outputs detections that would maximize the NAB score; i.e. it outputs only true positives at the beginning of each window. As discussed earlier the raw NAB scores from these two detectors are used to scale the NAB score for all other algorithms between 0 and 100. We also include a "random" detector that outputs a random anomaly score between 0 and 1 for each data instance. The score from this detector offers some intuition for the chance-level performance on NAB.

Each algorithm in NAB is allowed to output an anomaly score between 0 and 1. The score is then thresholded using a fixed threshold in order to detect an anomaly. NAB includes an automated hill-climbing search for selecting the optimal threshold for each algorithm. This search efficiently optimizes over the range of possible thresholds, where the objective function to be maximized is the NAB scoring function. The detection threshold is thus tuned based on the full NAB dataset, under the constraint that a single fixed number be used for all data files.

## IV. RESULTS

### A. Overall NAB Scores

Table 1 summarizes the scores for all algorithms on the three application profiles using the data files in NAB 1.0. The HTM detector achieves the best overall scores, followed by Etsy and Twitter. Although the HTM detector performs markedly better, the Etsy and Twitter algorithms perform significantly better than chance across all three application profiles.

TABLE I. NAB SCOREBOARD

| Detectors | Scores for Application Profiles[a] | | |
|---|---|---|---|
| | Standard | Reward low FP | Reward low FN |
| 1. Numenta HTM | 64.7 | 56.5 | 69.3 |
| 2. Twitter ADVec | 47.1 | 33.6 | 53.5 |
| 3. Etsy Skyline | 35.7 | 27.1 | 44.5 |
| 4. Random[b] | 16.8 | 5.8 | 25.9 |
| 5. Null | 0.0 | 0.0 | 0.0 |

[a.] Each algorithms' parameters were optimized to yield the best NAB scores possible.
[b.] Random detections do not yield scores ≈ 0 because of the NAB score optimization step.

### B. Results Analysis

A detailed analysis of the errors illustrates the primary strengths and weaknesses of each algorithm. Notice all algorithms dropped in score from the Standard to Reward low FP profiles, but why does the Skyline score decrease more than the others? For the NAB corpus of 365,558 data instances, Skyline detects 1161 as anomalous, whereas HTM and ADVec detect 387 and 612, respectively.[1] An algorithm that makes many detections is more likely to result in more FPs than one with fewer. Similarly, this algorithm would result in fewer FNs, as reflected in the Reward low FN results, where Skyline increased the most of the tested algorithms.

Investigating individual results files illustrates some of the interesting situations that arise in real time applications and how different detectors behave. Fig. 4 demonstrates the value of continuous learning. This file shows CPU usage on a production server over time and contains two anomalies. The first is a simple spike detected by all algorithms. The second is a sustained shift in the usage. Skyline and HTM both detect the change but then adapt to the new normal (with Skyline adapting quickest). Twitter ADVec however continues to generate anomalies for several days.

---
[1] To maximize the ADVec scores we tuned the "max_anom" and "period" parameters. The former was set to flag a maximum of 0.20% of the data instances in a given file as anomalous. The latter set the period length to 150 data instances (used during seasonal decomposition).

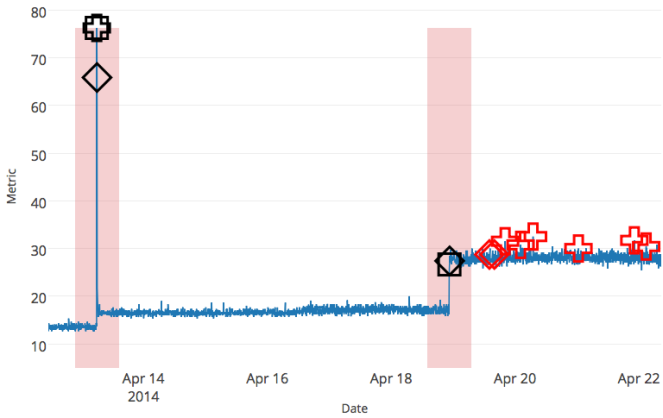

Fig. 4. Detection results for anomalies on a production server based on CPU usage. The shapes correspond to different detectors: HTM, Skyline, and ADVec are diamond, square, and plus respectively. For a given detector, the scored (i.e. the first) TP detection within each window is labeled in black. All FP detections are colored red. As in Fig. 2, the red shaded regions denote the anomaly windows.

Figs. 5 and 6 demonstrate temporal anomalies and their importance in early detection. Fig. 5 shows a close up of the results for the previously discussed machine temperature sensor data file. The first anomaly shown in this plot is a somewhat subtle temporal anomaly where the temporal behavior is unusual but individual readings are within the expected range. This anomaly (which preceded the catastrophic failure on February 8) is only detected by HTM. All three detectors detect the second anomaly, although Skyline and HTM detect it earlier than ADVec. In this plot HTM and Skyline also each have a false positive. Fig. 6 is another example demonstrating early detection. All three detectors detect the anomaly but HTM detects it three hours earlier due to a subtle shift in metric dynamics.

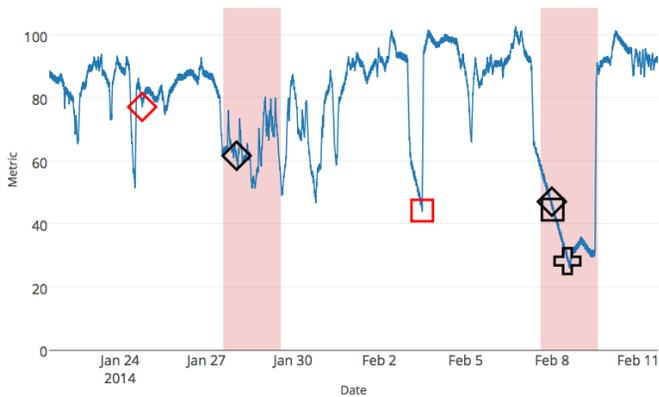

Fig. 5. A close up of the detection results for the machine temperature data of Figs. 1 (bottom) and 2. Only the HTM (diamond) detects the first anomaly (a purely temporal anomaly). All algorithms detect the second anomaly, HTM and Skyline hours before ADVec. HTM and Skyline (diamond) each show an FP.

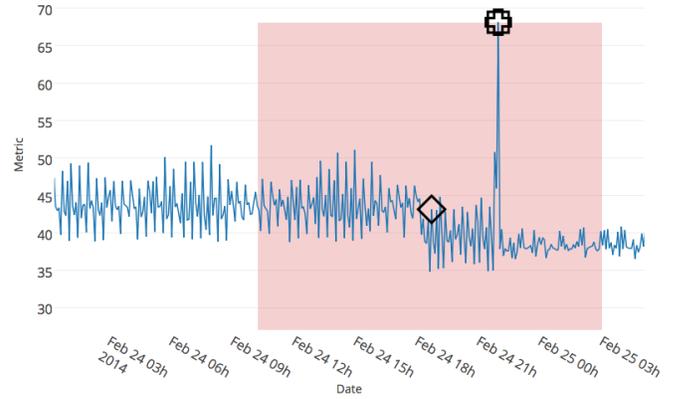

Fig. 6. Detection results demonstrating early detection. Although all detectors correctly identify the anomaly, the HTM (diamond) detects it three hours earlier than Skyline and ADVec due to a change in the metric dynamics.

Both figures illustrate a very common situation that occurs in practice. As demonstrated by the NAB data files, temporal changes in behavior often precede a large easily detectable shift. Temporal and sequence based anomaly detection techniques can thus detect anomalies in streaming data before they are easily visible. This provides hope that such algorithms can be used in production to provide early warning and help avoid catastrophes far more reliably than traditional techniques.

## V. CONCLUSION AND FUTURE WORK

The aim of NAB is to provide the anomaly detection research community with a controlled and repeatable environment of tools to test and measure different algorithms for real-time streaming applications. The contributions of this work are threefold:

I. Benchmark dataset: real world time-series data files from a variety of domains, labeled with anomalies. Easily accessible real data for streaming applications is rare, and its value for algorithm practitioners is significant. In this paper we have noted some of the insights such as the presence of subtle changes in dynamics preceding large-scale failures.

II. Performance evaluation: a scoring philosophy designed for real time applications. The scoring system in NAB augments traditional metrics by incorporating time, explicitly rewarding early detection. The addition of easily tuned application profiles allows developers to explicitly test algorithms for their specific application requirements.

III. Code library: fully open source repository complete with data, algorithms, and documentation.

We discussed the NAB components and methods, and presented evaluation results for several open source detection algorithms. Although the HTM algorithm outperforms the other tested anomaly detectors, the results show there is still room for improvement. Investigating the results has allowed us to pinpoint the strengths and shortcomings of all algorithms.

We have found NAB to be a comprehensive, robust evaluation tool for real-world anomaly detection algorithms.

NAB v1.0 is ready for researchers to evalute their algorithms and report their results. To move beyond NAB v1.0, future work on NAB will involve adding more real-world data files to the corpus. We also anticipate incorporating multivariate anomaly detection and categorical data. Over time we hope researchers can use NAB to test and develop a large number of anomaly detection algorithms for the specific purpose of applying them to real time streaming applications.

ACKNOWLEDGMENT

We would like to thank Jeff Hawkins, Ian Danforth, Celeste Baranski, Jay Gokhale, and Tom Silver for their contributions to this paper. We also thank the IEEE reviewers for comments that helped improve overall readability.